\pdfoutput=1

\documentclass[11pt]{article}

\usepackage[acceptedWithA]{tacl2018v2}

\usepackage{times}
\usepackage{latexsym}
\usepackage{enumitem}

\usepackage[T1]{fontenc}

\usepackage[utf8]{inputenc}

\usepackage{microtype}

\usepackage{color}
\usepackage[flushleft]{threeparttable}
\usepackage{booktabs}
\usepackage{graphicx}
\usepackage{multirow}
\usepackage{makecell}
\usepackage{xspace}
\usepackage{amssymb}
\usepackage{amsmath}
\usepackage{arydshln}
\usepackage{gensymb}

\usepackage{pifont}

\usepackage{lipsum}
\newcommand{\sep}{\texttt{[SEP]}\xspace}
\newcommand{\blk}{\texttt{[BLK]}\xspace}
\newcommand{\cls}{\texttt{[CLS]}\xspace}

\newcommand{\vilah}{H-VILA\xspace}
\newcommand{\vilai}{I-VILA\xspace}
\newcommand{\sshard}{\textsc{S2-VL}\xspace}  
\newcommand{\dataset}{\textsc{S2-VL}\xspace} 
\newcommand{\benchmark}{\textsc{S2-VLUE}\xspace} %
\newcommand{\mfscore}{Macro F1\xspace}
\newcommand{\assumption}{group uniformity assumption\xspace}

\title{VILA: Improving Structured Content Extraction from Scientific PDFs Using Visual Layout Groups}

\author{Zejiang Shen$^1$ \quad 
        Kyle Lo$^1$ \quad 
        Lucy Lu Wang$^1$ \quad 
        Bailey Kuehl$^1$ \quad \\
        {\bf
        Daniel S. Weld$^{1,2}$ \quad
        Doug Downey$^{1,3}$
        }
        \\ [1mm]
        $^1$Allen Institute for AI \quad $^2$University of Washington \quad
        $^3$Northwestern University
        \\
        {\tt\small \{shannons, kylel, lucyw, baileyk, danw, dougd\}@allenai.org}
}

\begin{document}

\maketitle

\begin{abstract}
  Accurately extracting structured content from PDFs is a critical first step for NLP over scientific papers.
  Recent work has improved extraction accuracy by incorporating elementary layout information, e.g., each token's 2D position on the page, into language model pretraining.
  We introduce new methods that explicitly model VIsual LAyout (VILA) \emph{groups}, i.e., text lines or text blocks, to further improve performance.
  In our \vilai approach, we show that simply inserting special tokens denoting layout group boundaries into model inputs can lead to a 1.9\% \mfscore improvement in token classification.
  In the \vilah approach, we show that hierarchical encoding of layout-groups can result in up-to 47\% inference time reduction with less than 0.8\% \mfscore loss.
  Unlike prior layout-aware approaches, our methods do not require expensive additional pretraining, only fine-tuning, which we show can reduce training cost by up to 95\%.
  Experiments are conducted on a newly curated evaluation suite, \benchmark, that unifies existing automatically-labeled datasets and includes a new dataset of manual annotations covering diverse papers from 19 scientific disciplines.
  Pre-trained weights, benchmark datasets, and source code are available at \url{https://github.com/allenai/VILA}.
\end{abstract}

\section{Introduction}

Scientific papers are usually distributed in Portable Document Format (PDF) without extensive semantic markup. 
Extracting structured document representations from these PDF files---i.e., identifying title and author blocks, figures, references, and so on---is a critical first step for downstream NLP tasks~\cite{beltagy-etal-2019-scibert, wang2020cord} and is important for improving PDF accessibility~\cite{wang2021improving}. 

\begin{figure}[t]
    \centering
    \includegraphics[width=\linewidth]{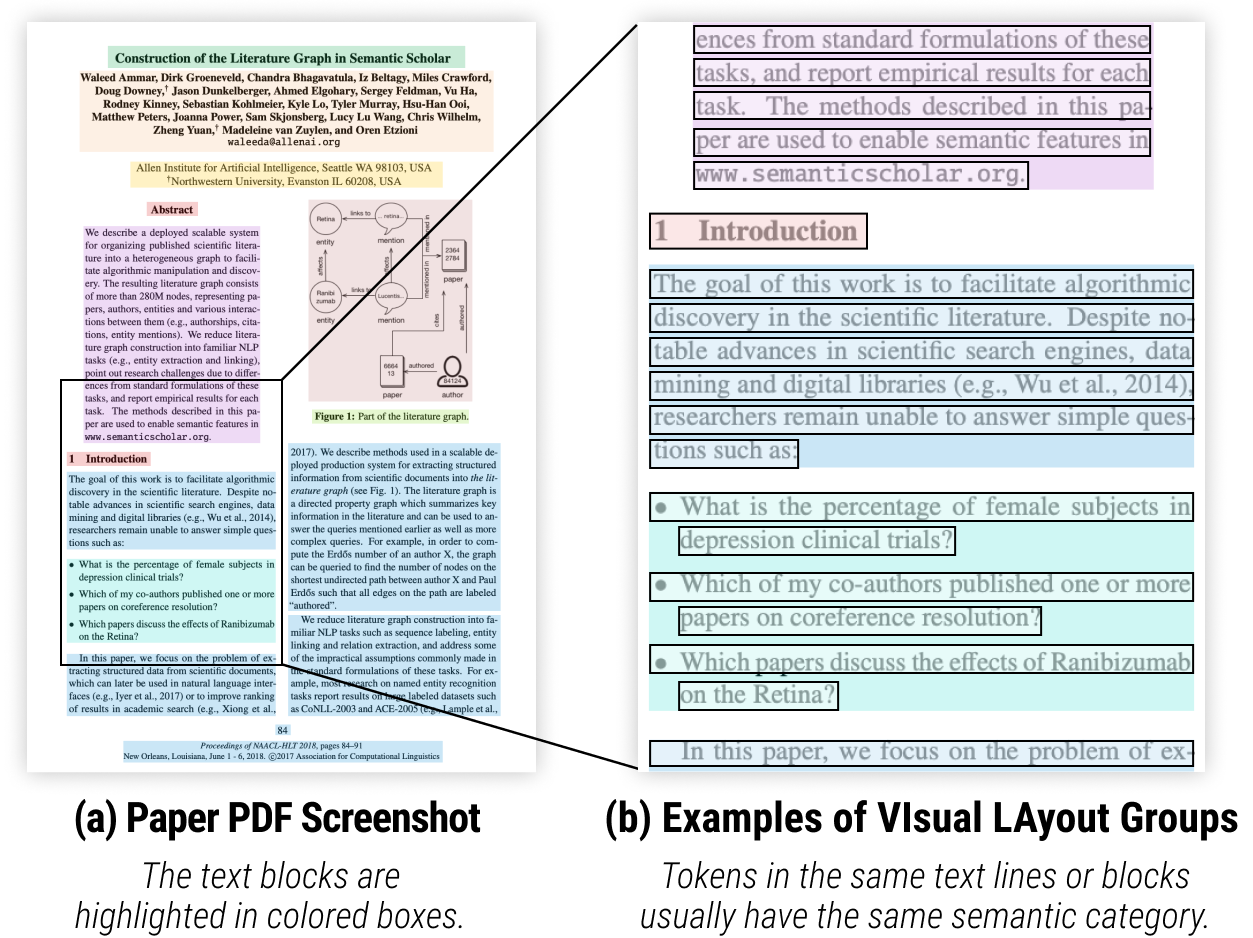}
    \caption{(a) Real-world scientific documents often have intricate layout structures, so analyzing only flattened raw text forfeits valuable information, yielding sub-optimal results. (b) The complex structures can be broken down into groups (text blocks or lines) that are composed of tokens with the same semantic category.}
    \label{fig:teaser}
\end{figure}

Recent work demonstrates that document layout information can be used to enhance content extraction via large-scale, layout-aware pretraining~\cite{xu2020layoutlm, xu2020layoutlmv2, li2021selfdoc}. 
However, these methods only consider individual tokens' 2D positions and do not explicitly model high-level layout structures like the grouping of text into lines and blocks (see Figure \ref{fig:teaser} for example), limiting accuracy. %
Further, existing methods come with enormous computational costs: they rely on further pretraining an existing pretrained model like BERT~\citep{devlin-etal-2019-bert} on layout-enriched input, and achieving the best performance from the models requires more than a thousand~\citep{xu2020layoutlm} to several thousand~\citep{xu2020layoutlmv2} GPU-hours. 
This means swapping in a new pretrained text model or experimenting with new layout-aware architectures can be prohibitively expensive, incompatible with the goals of green AI \cite{schwartz2020green}.  

In this paper, we explore how to improve the accuracy and efficiency of structured content extraction from scientific documents by using \textbf{VI}sual \textbf{LA}yout (VILA) groups. 
Following \citet{zhong2019publaynet} and \citet{tkaczyk2015cermine}, our methods use the idea that a document page can be segmented into visual groups of tokens (either lines or blocks), and that the tokens within each group generally have the same semantic category, which we refer to as the \emph{\assumption} (see Figure~\ref{fig:teaser}(b)).
Given text lines or blocks generated by rule-based PDF parsers~\cite{tkaczyk2015cermine} or vision models~\cite{zhong2019publaynet}, we design two different methods to incorporate the VILA groups and the assumption into modeling:
the \textbf{\vilai} model adds layout \textbf{i}ndicator tokens to textual inputs to improve the accuracy of existing BERT-based language models,
while the \textbf{\vilah} model uses VILA structures to define a \textbf{h}ierarchical model that models pages as collections of groups rather than of individual tokens, increasing inference efficiency.

Previous datasets for evaluating PDF content extraction rely on machine-generated labels of imperfect quality, and comprise papers from a limited range of scientific disciplines. 
To better evaluate our proposed methods, we design a new benchmark suite, Semantic Scholar Visual Layout-enhanced Scientific Text Understanding Evaluation (\textbf{\benchmark}).
The benchmark extends two existing resources~\cite{tkaczyk2015cermine,li2020docbank} and introduces a newly curated dataset, \sshard, which contains high-quality human annotations for papers across 19 disciplines. 

Our contributions are as follows:
\begin{enumerate}[noitemsep]
  \item We introduce a new strategy for PDF content extraction that uses VILA structures to inject layout information into language models, and show that this improves accuracy {\em without} the expensive pretraining required by existing methods, and generalizes to different language models.
  \item We design two models that incorporate VILA features differently. The \vilai model injects layout indicator tokens into the input texts and improves prediction accuracy (up to +1.9\% \mfscore) and consistency compared to the previous layout-augmented language model LayoutLM~\cite{xu2020layoutlm}. The \vilah model performs group-level predictions and can reduce model inference time by 47\% with less than 0.8\% loss in \mfscore. 
  \item We construct a unified benchmark suite \benchmark which enhances existing datasets with VILA structures, and introduce a novel dataset \dataset that addresses gaps in existing resources. \dataset contains hand-annotated gold labels for 15 token categories on papers spanning 19 disciplines. 
\end{enumerate}
The benchmark datasets, modeling code, and trained weights are available at \url{https://github.com/allenai/VILA}.

\section{Related Work}

\subsection{Structured Content Extraction for Scientific Documents}

Prior work on structured content extraction for scientific documents usually relies on textual or visual features. 
Text-based methods like ScienceParse~\cite{ammar-etal-2018-construction}, GROBID~\cite{GROBID} or Corpus Conversion Service~\cite{staar2018corpus} combine PDF-to-text parsing engines like CERMINE~\cite{tkaczyk2015cermine} or pdfalto,\footnote{\url{https://github.com/kermitt2/pdfalto} (last accessed Jan. 1, 2022).} which output a sequence of tokens extracted from a PDF, with machine learning models like RNN~\cite{hochreiter1997long}, CRF~\cite{lafferty2001conditional}, or Random Forest~\cite{breiman2001random} trained to classify the token categories of the sequence. 
Though these models are practical and fairly efficient, they fall short in prediction accuracy or generalize poorly to out-of-domain documents. 
Vision-based Approaches~\cite{zhong2019publaynet,he2017mask,siegel2018extracting}, on the other hand, treat the parsing task as an image object detection problem: given document images, the models predict rectangular bounding boxes, segmenting the page into individual components of different categories.
These models excel at capturing complex visual layout structures like figures or tables, but because they operate only on visual signals without textual information, they cannot accurately predict fine-grained semantic categories like title, author, or abstract, which are of central importance for parsing scientific documents.

\subsection{Layout-aware Language Models}

Recent methods on layout-aware language models improve prediction accuracy by jointly modeling documents' textual and visual signals. 
LayoutLM~\cite{xu2020layoutlm} learns a set of novel positional embeddings that can encode tokens' 2D spatial location on the page and improves accuracy on scientific document parsing~\cite{li2020docbank}. 
More recent work~\cite{xu2020layoutlmv2, li2021selfdoc} aims to encode the document in a multimodal fashion by modeling text and images together. 
However, these existing joint-approach models require expensive pretraining, and may be less efficient as a consequence of their joint inputs~\citep{xu2020layoutlmv2},
making them less suitable for deployment at scale. 
In this work, we aim to incorporate document layout features in the form of visual layout groupings, in novel ways that improve or match performance without the need for expensive pretraining. 
Our work is well-aligned with recent efforts for incorporating structural information into language models~\cite{lee-etal-2020-slm,bai2020segatron,yang2020beyond,zhang-etal-2019-hibert}. 

\begin{figure}[t]
    \centering
    \includegraphics[width=\linewidth]{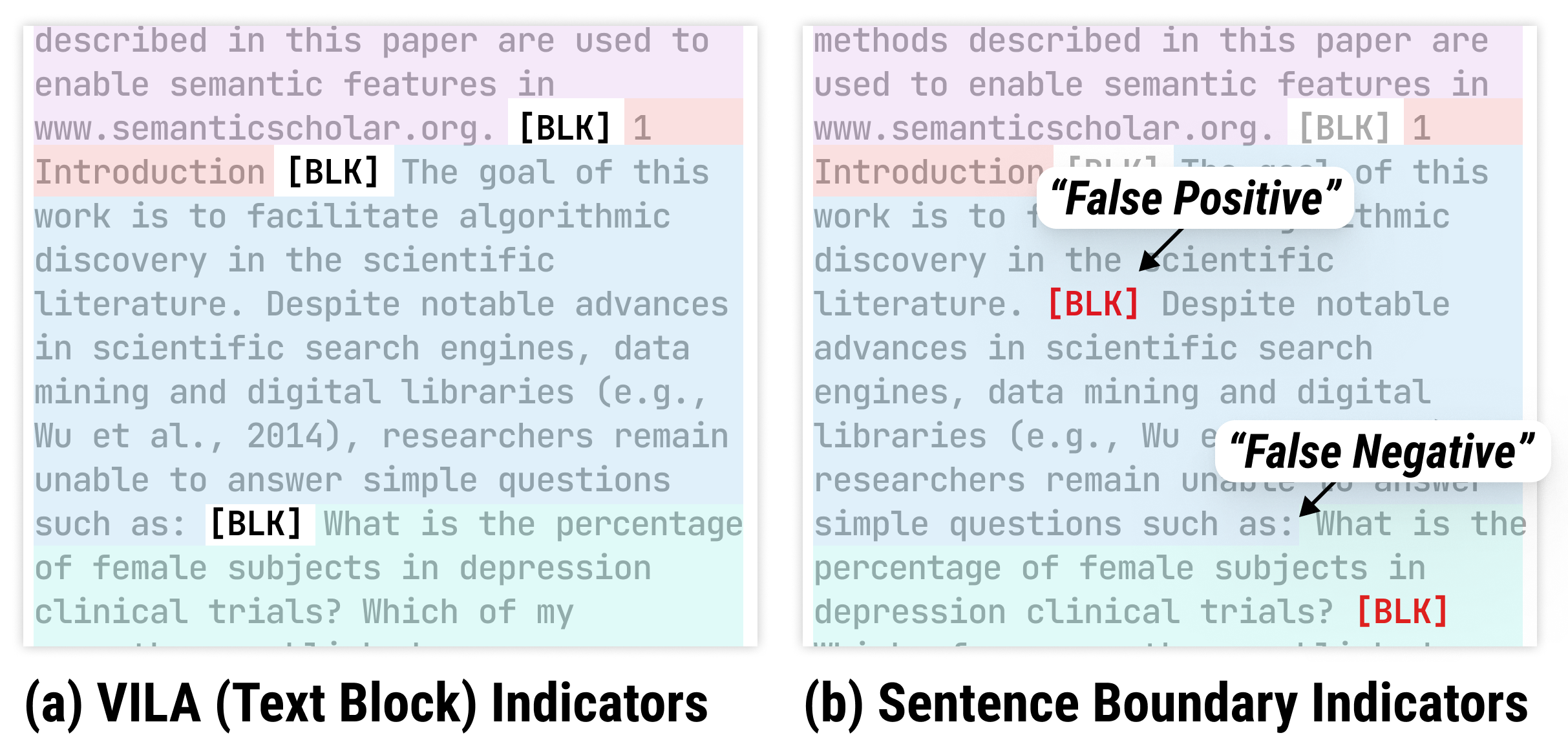}
    \caption{Comparing inserting indicator tokens \blk based on VILA groups and sentence boundaries. Indicators representing VILA groups  (e.g., text blocks in the left figure) are usually consistent with the token category changes (illustrated by the background color in (a)), while sentence boundary indicators fail to provide helpful hints (both "false positive"s and "false negative"s occur frequently in (b)). Best viewed in color.} 
    \label{fig:ivila}
    \vspace{-2mm}
  \end{figure}
\subsection{Training and Evaluation Datasets}
The available training and evaluation datasets for scientific content extraction models are automatically generated from author-provided source data, e.g. GROTOAP2 \citep{tkaczyk2014grotoap2} and PubLayNet \citep{zhong2019publaynet} are constructed from PubMed Central XML
and DocBank \citep{li2020docbank} from arXiv LaTeX source.
Despite their large sample sizes, these datasets have limited layout variation, leading to poor generalization to papers from other disciplines with distinct layouts. 
Also, due to the heuristic nature in which the data are automatically labeled, they contain systematic classification errors that can affect downstream modeling performance.
We elaborate on the limitations of GROTOAP2~\cite{tkaczyk2014grotoap2} and DocBank~\cite{li2020docbank} in Section~\ref{sec:datasets}. 
PubLayNet~\cite{zhong2019publaynet} provides high-quality text block annotations on 330k document pages, but its annotations only cover five distinct categories.
\citet{livathinos2021robust} and \citet{staar2018corpus} curated a multi-disciplinary, manually annotated dataset of 2,940 paper pages, but only make available the processed page features without the raw text or source PDFs needed for experiments with layout-aware methods.
We introduce a new evaluation dataset, \dataset, to address limitations in these existing datasets.

\begin{figure*}[t]
    \centering
    \includegraphics[width=0.7\linewidth]{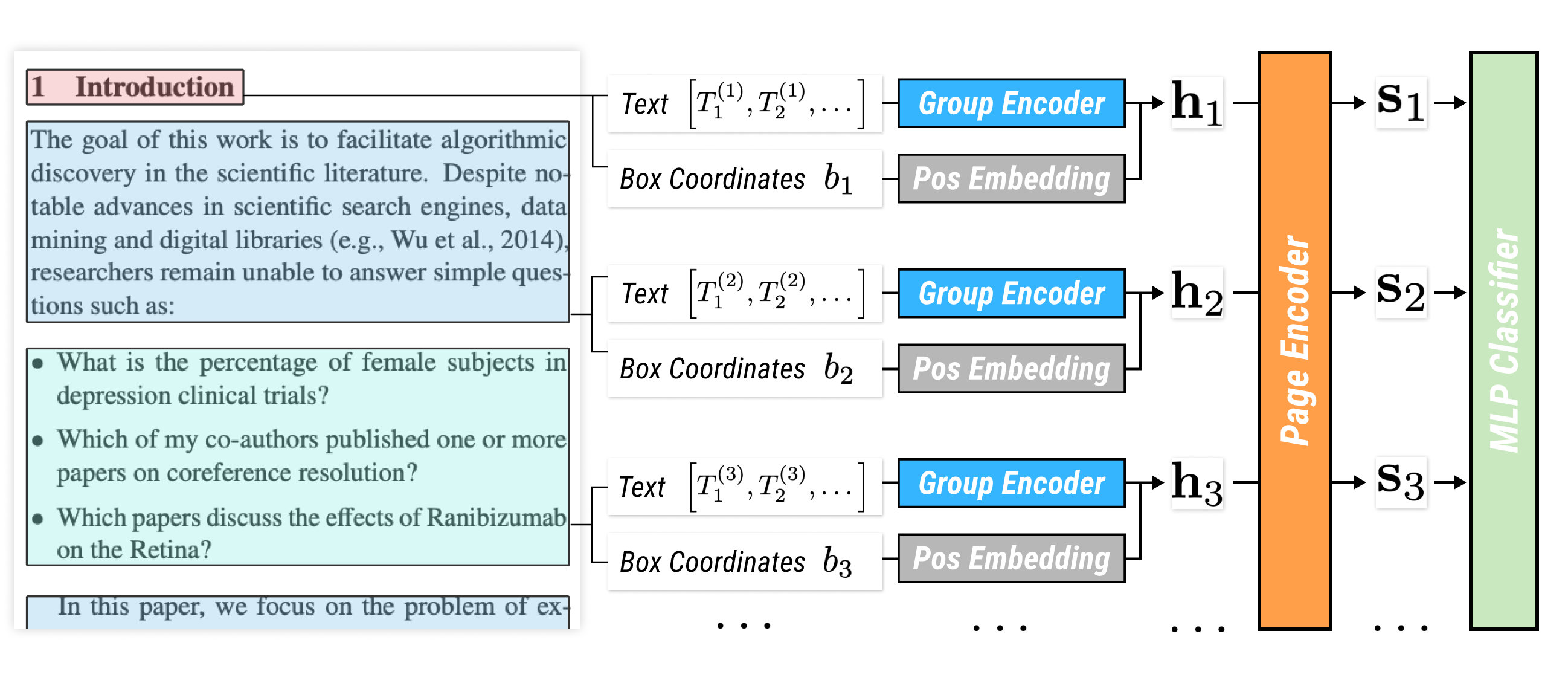}
    \caption{Illustration of the \vilah model. Texts from each visual layout group are encoded separatedly using the group encoder, and the generated representation are subsequently modeled by a page encoder. The semantic category are predicted at the group-level, which significantly improves efficiency. } 
    \label{fig:hvila}
    \vspace{-3mm}
\end{figure*}

\section{Methods}

\subsection{Problem Formulation}
\label{sec:formulation}

Following prior work~\cite{tkaczyk2015cermine,li2020docbank}, our task is to map each token $t_i$ in an input sequence $T=(t_1, \dots, t_n)$ to its semantic category $c_i$ (e.g. title, body text, reference, etc.).
Input tokens are extracted via PDF-to-text tools, which output both the word $t_i$ and its 2D position on the page, a rectangular bounding box $a_i=(x_0, y_0, x_1, y_1)$ denoting the left, top, right, and bottom coordinate of the word. 
The order of tokens in sequence $T$ may not reflect the actual reading order of the text due to errors in PDF-to-text conversion (e.g., in the original DocBank dataset~\cite{li2020docbank}), which poses an additional challenge to language models pre-trained on regular texts.

Besides the token sequence $T$, additional visual structures $G$ can also be retrieved from the source document. 
Scientific papers are organized into {\em groups} of tokens (lines or blocks), which consist of consecutive pieces of text that can be segmented from other pieces based on spatial gaps. 
The group information 
can be extracted via visual layout detection models~\cite{zhong2019publaynet,he2017mask} or rule-based PDF parsing~\cite{tkaczyk2015cermine}. 

Formally, given an input page, the group detector identifies a series of $m$ rectangular boxes for each group $b_j\in B=\{b_1, \dots, b_m\}$ in the input document page, where $b_j=(x_0, y_0, x_1, y_1)$ denotes the box coordinates.  
Page tokens are allocated to the visual groups $g_j = (b_j, T^{(j)})$, where $T^{(j)}=\{t_i \mid a_i\preceq b_j, t_i\in T \}$ contains all tokens in the $j$-th group, and $a_i\preceq b_j$ denotes that the center point of token $t_i$'s bounding box $a_i$ is strictly within the group box $b_i$. 
When two group regions overlap and share common tokens, the system assigns the common tokens to the earlier group by the estimated reading order from the PDF parser. 
We refer to text block groups of a page as $G^{(\mathcal{B})}$ and text line groups as $G^{(\mathcal{L})}$. 
In our case, we define text lines as consecutive tokens appearing at the nearly same vertical position.\footnote{Or horizontal position, when the text is written vertically.}
Text blocks are sets of adjacent text lines with gaps smaller than a certain threshold, and ideally the same semantic category.  That is, even two close lines of different semantic categories should be allocated to separate blocks, and in our models we use a block detector trained toward this objective.  In practice, block or line detectors may generate incorrect predictions.

In the following sections, we describe our two models, \vilai and \vilah.  The models take a BERT-based pretrained language model as a foundation, which may or may not itself be layout-aware (we experiment with DistilBERT, BERT, RoBERTa, and LayoutLM in our experiments).  Our models then augment the base model to incorporate group structures, as detailed below.

\begin{figure*}[t]
    \centering
    \includegraphics[width=\linewidth]{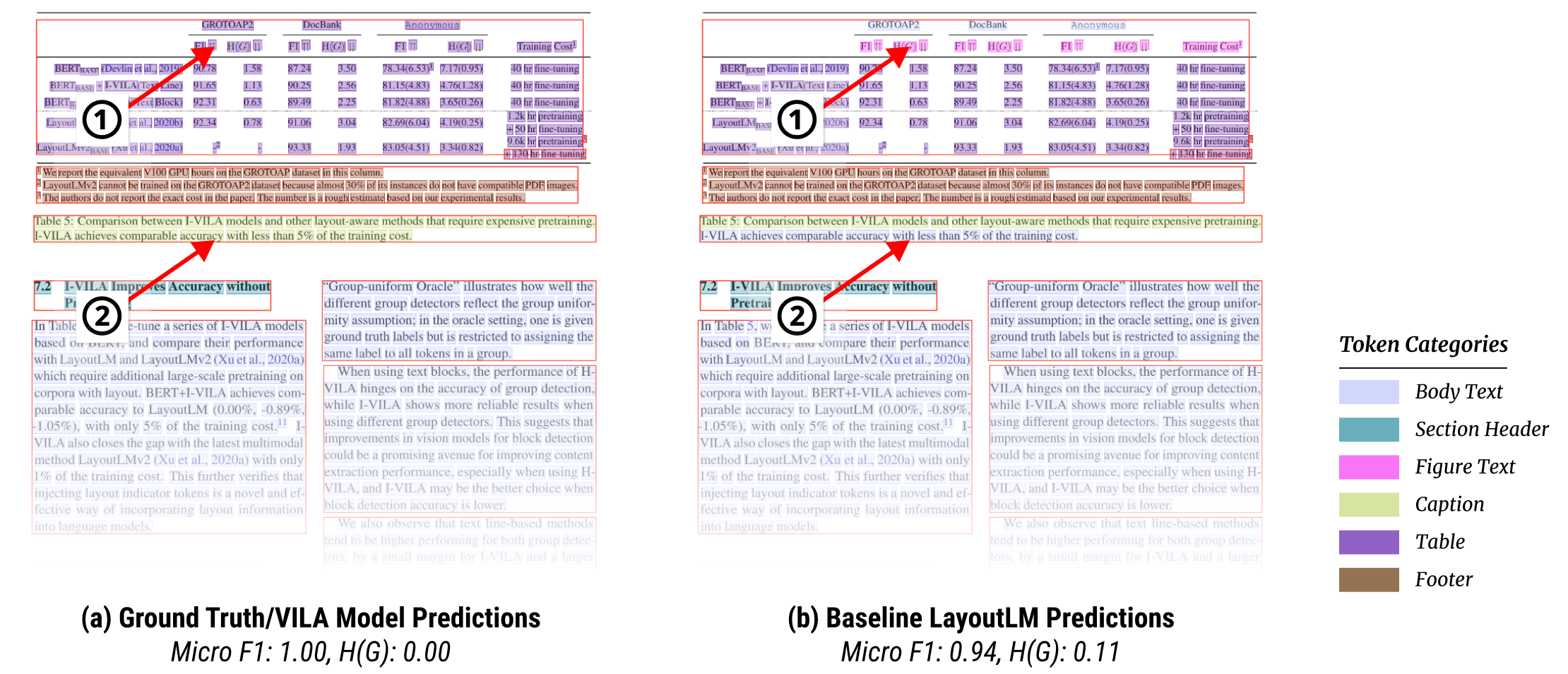}
    \caption{Model predictions for the 10th page of our paper draft.  We present the token category and text block bounding boxes (highlighted in red rectangles) based on the (a) ground-truth annotations and model predictions from both \vilai and \vilah models (the three results happen to be identical) and (b) model predictions from the LayoutLM model. When VILA is injected, the model achieves more consistent predictions for the example, as indicated by arrows (1) and (2) in the figure. Best view in color.}
    \label{fig:pred}
    \vspace{-3mm}
  \end{figure*}

\subsection{\vilai: Injecting Visual Layout Indicators}

According to the \assumption, token categories are homogeneous within a group, and categorical changes should happen at group boundaries. 
This suggests that layout information should be incorporated in a way that informs token \emph{category consistency} intra-group and signals possible token \emph{category changes} inter-group. 

Our first method supplies VILA structures by inserting a special layout indicator token at each group boundary in the input text, and models this with a pretrained language model (which may or may not be position-aware).  We refer to this as the \vilai method. 
As shown in Figure~\ref{fig:ivila}(a), the inserted tokens partition the text into segments that provide helpful structure to the model, hinting at possible category changes. 
In \vilai, the special tokens are seen at all layers of the model, providing VILA signals at different stages of modeling, rather than only providing positional information at the initial embedding layers as in LayoutLM \cite{xu2020layoutlm}. 
We empirically show that BERT-based models can learn to leverage such special tokens to improve both the accuracy and the consistency of category predictions, even without an additional loss penalizing inconsistent intra-group predictions.

In practice, given $G$, we linearize tokens $T^{(j)}$ from each group and flatten them into a 1D sequence. 
To avoid capturing confounding information in existing pretraining tasks, we insert a new token previously unseen by the model, \blk, in-between text from different groups $T^{(j)}$.
The resulting input sequence is of the form $[\cls, T^{(1)}_1, \dots, T^{(j)}_{n_j}, \blk, T^{(j+1)}_{1}, \dots,$ $ T^{(m)}_{n_m}, \sep ]$, where $T^{(j)}_{i}$ and $n_j$ indicate the $i$-th token and the total number of tokens respectively in the $j$-th group, and \cls and \sep are the special tokens used by the BERT model and are inserted to preserve a similar input structure.\footnote{The \cls and \sep tokens are only inserted at the beginning or end of each input sequence, and they do not represent the sentence boundaries in this case.}
The 
BERT-based models are fine-tuned on the token classification objective with a cross entropy loss.
When \vilai uses a visual pretrained language model as input, such as LayoutLM \cite{xu2020layoutlm}, the positional embeddings for the newly injected \blk tokens are generated from the corresponding group's bounding box $b_j$.

\subsection{\vilah: Visual Layout-guided Hierarchical Model}
\label{sec:vila-hie}

The uniformity of group token categories also suggests the possibility of building a group-level classifier. 
Inspired by recent advances in modeling long documents, hierarchical structures~\cite{yang2020beyond,zhang-etal-2019-hibert} provide an ideal architecture for the end task while optimizing for computational cost. 
Illustrated in  Figure~\ref{fig:hvila}, our hierarchical approach uses two transformer-based models, one to encode each group in terms of its words, and another modeling the whole document in terms of the groups.  We provide the details below.  

\textbf{The Group Encoder} is a $l_{\text{g}}$-layer transformer that converts each group $g_i$ into a hidden vector $\mathbf{h}_i$. 
Following the typical transformer model setting~\cite{transformer}, the model takes a sequence of tokens $T^{(j)}$ within a group as input, and maps each token $T^{(j)}_i$ into a dense vector $\mathbf{e}^{(j)}_{i}$ of dimension $d$. Subsequently, a group vector aggregation function $f: R^{n_j\times d} \to R^d$ is applied that projects the token representations $\left(\mathbf{e}^{(j)}_1, \dots, \mathbf{e}^{(j)}_{n_j}\right)$ to a single vector $\tilde{\mathbf{h}}_j$ that represents the group's textual information.
A group's 2D spatial information is incorporated in the form of positional embeddings, and the final group representation $\mathbf{h}_j$ can be calculated as: 
\begin{equation}
\mathbf{h}_j = \tilde{\mathbf{h}}_j + p(b_j) . 
\end{equation}
where $p$ is the 2D positional embedding similar to the one used in LayoutLM:
\begin{align}
  p(b) = & E_x(x_0) + E_x(x_1) + E_w(x_1-x_0) + \\ \nonumber
         & E_y(y_0) + E_y(y_1) + E_h(y_1-y_0) ,
\end{align}
where $E_x$, $E_x$, $E_w$, $E_h$ are the embedding matrices for x, y coordinates and width and height.
In practice, we find that injecting positional information using the bounding box of the first token within the group leads to better results, and we choose group vector aggregation function $f$ to be the average over all tokens representations. 

\textbf{The Page Encoder} is another stacked transformer model of $l_{\text{p}}$ layers that operates on the group representation $\mathbf{h}_j$ generated by the group encoder. 
It generates a final group representation $\mathbf{s}_j$ for downstream classification. A MLP-based linear classifier is attached thereafter, and is trained to generate the group-level category probability $p_{jc}$. 

Different from previous work~\cite{yang2020beyond}, we restrict the choice of $l_{\text{g}}$ and $l_{\text{p}}$ to $\{1, 12\}$ such that we can load pre-trained weights from BERT base models. 
Therefore, no additional pretraining is required, and the \vilah model can be fine-tuned directly for the downstream classification task. 
Specifically, we set $l_{\text{g}}=1$ and initialize the group encoder from the first-layer transformer weights of BERT. 
The page encoder is configured as either a one-layer transformer or a 12-layer transformer that resembles a full LayoutLM model. 
Weights are initialized from the first-layer or full 12 layers of the LayoutLM model, which is trained to model texts in conjunction with their positions. 

\textbf{Group Token Truncation} As suggested in \citet{yang2020beyond}'s work, when an input document of length $N$ is evenly split into segments of $L_s$, the memory footprint of the hierarchical model is $O(l_{\text{g}}NL_s +l_{\text{p}}(\frac{N}{L_s})^2)$, and for long documents with $N\gg L_s$, it approximates as $O(N^2/L_s^2)$. 
However, in our case, it is infeasible to adopt the Greedy Sentence Filling technique~\cite{yang2020beyond} as it mingles signals from different groups and obfuscates group structures.
It is also less desirable to simply use the maximum token count per group $\max_{1\le j \le m}{n_j}$ to batch the contents due to the high variance of group token length (see Table~\ref{table:dataset}).
Instead, we choose a group token truncation count $\tilde{n}$ empirically based on statistics of the group token length distribution such that $N\approx \tilde{n}m$, and use the first $\tilde{n}$ to aggregate the group hidden vector $\mathbf{h}_j$ for all groups (we pad the sequence to $\tilde{n}$ when it is shorter). 

\begin{table*}[th]
    \centering
    \resizebox{0.9\linewidth}{!}{
    \begin{threeparttable}
        \begin{tabular}{l|l|l|l}
        \toprule
        & \textbf{GROTOAP2}                         & \textbf{DocBank}    & \textbf{\dataset}\\
        \midrule
        Train / Dev / Test Pages& 83k / 18k / 18k  & 398k / 50k / 50k    & 1.3k\tnote{1}             \\
        Annotation Method    & Automatic        & Automatic           & Human Annotation \\ 
        Scientific Discipline & Life Science          & Math / Physics / CS & 19 Disciplines       \\
        Visual Layout Group   & PDF parsing           & Vision model        & Gold Label / Detection methods \\
        Number of Categories      & 22 & 12 & 15 \\
        \midrule
        Average Token Count\tnote{2} & 1203 (591) & 838 (503)     & 790 (453) \\
        Average Text Line Count      & 90 (51)    & 60 (34)       & 64 (54) \\
        Average Text Block Count     & 12 (16)    & 15 (8)        & 22 (36) \\
        \bottomrule
        \end{tabular}
        \begin{tablenotes}
            \small
            \item[1] This is the total number of pages in the \dataset dataset; we use 5-fold cross-validation for training and testing. 
            \item[2] We report the average token, text line, and text block count per page, with standard deviations in parentheses. 
        \end{tablenotes}
    \end{threeparttable}
    }
\caption{Details for the three datasets in the \benchmark benchmark.%
}
\label{table:dataset}
\end{table*}

\section{Benchmark Suite}

\label{sec:datasets}

To systematically evaluate the proposed methods, we develop the the Semantic Scholar \textbf{V}isual \textbf{L}ayout-enhanced Scientific Text \textbf{U}nderstanding \textbf{E}valuation (\benchmark) benchmark suite. 
\benchmark consists of three datasets---two previously released resources which we augment with VILA information, and a new hand-curated dataset \dataset.

Key statistics for \benchmark are provided in Table~\ref{table:dataset}. 
Notably, the three constituent datasets differ with respect to their: 1) annotation method, 2) VILA generation method, and 3) paper domain coverage. 
We provide details below. 

\paragraph{GROTOAP2} The GROTOAP2 dataset~\cite{tkaczyk2014grotoap2} is automatically annotated.
Its text block and line groupings come from the CERMINE PDF parsing tool~\cite{tkaczyk2015cermine}; text block category labels are then obtained by pairing block texts with structured data from document source files obtained from PubMed Central. 
A small subset of data is inspected by experts, and a set of post-processing heuristics is developed to further improve annotation quality. 
Since token categories are annotated by group, the dataset achieves perfect accordance between token labels and VILA structures. 
However, the method of rule-based PDF parsing employed by the authors introduces labeling inaccuracies due to imperfect VILA detection: the authors find that block-level annotation accuracy achieves only 92 \mfscore in a small gold evaluation set. 
Additionally, all samples are extracted from the PMC Open Access Subset\footnote{\url{https://www.ncbi.nlm.nih.gov/pmc/tools/openftlist/} (last accessed Jan. 1, 2022).} that includes only life sciences publications; these papers have less representation of classification types like ``equation'', which are common in other scientific disciplines. 

\paragraph{DocBank} The DocBank dataset~\cite{li2020docbank} is fully machine-labeled without any postprocessing heuristics or human assessment.  
The authors first identify token categories by automatically parsing the source TEX files available from arXiv. 
Text block annotations are then generated by grouping together tokens of the same category using connected component analysis. 
However, only a specific set of token tags is extracted from the main TEX file for each paper, leading to inaccurate and incomplete token labels, especially for papers employing LaTeX macro commands,\footnote{For example, in DocBank, ``Figure 1'' in a figure caption block is usually labeled as ``paragraph'' rather than ``caption''. DocBank labels all tokens that are not explicitly contained in the set of processed LaTeX tags as ``paragraph.''} and thus, incorrect visual groupings. %
Hence, we develop a Mask R-CNN-based vision layout detection model based on a collection of existing resources~\cite{zhong2019publaynet,MFDDataset, he2017mask,shen2021layoutparser} to fix these inaccuracies and generate trustworthy VILA annotations at both the text block and line level.\footnote{
The original generation method for DocBank requires rendering LaTeX source, which results in layouts different from the publicly available versions of these documents on arXiv. However, because the authors of the dataset only provide document page images, rather than the rendered PDF, we can only use image-based approaches for layout detection. We refer readers to the appendix for details.}
As a result, this dataset can be used to evaluate VILA models under a different setting, since the VILA structures are generated independently from the token annotations. 
Because the papers in DocBank are from arXiv, however, they primarily represent domains like Computer Science, Physics, and Mathematics, limiting the amount of layout variation.

\paragraph{\dataset} 
We introduce a new dataset to address the three major drawbacks in existing work: 1) annotation quality, 2) VILA fidelity, and 3) domain coverage. 
\dataset is manually labeled by graduate students who frequently read scientific papers. 
Using the PAWLS annotation tool~\cite{neumann2021pawls}, annotators draw rectangular text blocks directly on each PDF page, and specify the block-level semantic categories from 15 possible candidates.\footnote{Of our defined categories, 12 are common fields and taken directly from other similar datasets, e.g., title, abstract etc. We add three categories: equation, header, and footer, which commonly occur in scientific papers and are included in full text mining resources like S2ORC~\cite{lo-wang-2020-s2orc} and CORD-19~\cite{wang2020cord}.}
Tokens within a group can therefore inherit the category from the parent text block. 
Inter-annotator agreement, in terms of token-level accuracy measured on a 12-paper subset, is high at 0.95.
The ground-truth VILA labels in \dataset can be used to fine-tune visual layout detection models, and paper PDFs are also included, making PDF-based structure parsing feasible: this enables VILA annotations to be created by different means, which is helpful for benchmarking new VILA-based models.  
Moreover, \dataset currently contains 1337 pages of 87 papers from 19 different disciplines, including e.g. Philosophy and Sociology which are not present in previous data sets. 

Overall, the datasets in \benchmark cover a wide range of academic disciplines with different layouts. 
The VILA structures in the three component datasets are curated differently, which helps to evaluate the generality of VILA-based methods.

\begin{table*}[htbp]
    \centering
    \resizebox{0.95\linewidth}{!}{
    \begin{threeparttable}
        \renewcommand{\arraystretch}{1.35}%

        \begin{tabular}{@{}rrrrcrrcrrcr@{}}
            \toprule
            & \multicolumn{2}{c}{GROTOAP2} & \phantom{\small{1}} & \multicolumn{2}{c}{DocBank}  & \phantom{\small{1}} & \multicolumn{2}{c}{\dataset\tnote{1}}  \\
            \cmidrule{2-3} \cmidrule{5-6} \cmidrule{8-9}
            & \mfscore $\upuparrows$ & $\mathrm{H}(G)$ $\downdownarrows$ &  & \mfscore $\upuparrows$ & $\mathrm{H}(G)$ $\downdownarrows$ & & \mfscore $\upuparrows$ & $\mathrm{H}(G)$ $\downdownarrows$ \\
            \midrule
            $\text{LayoutLM}_{\text{BASE}}$~\cite{xu2020layoutlm}        &  92.34          & 0.78          &  & 91.06          & 2.64          & & 82.69(6.04) & 4.19(0.25)   \\
            
            $\text{LayoutLM}_{\text{BASE}}$ + Sentence Breaks        &  91.83          & 0.78          &  & 91.44          & 2.62          & & 82.81(5.21)          & 4.21(0.55)        \\
            \cdashline{1-12}[.4pt/1pt]
            $\text{LayoutLM}_{\text{BASE}}$ + \textbf{\vilai}(Text Line)    &  92.37          & 0.73          &  & \textbf{92.79} & 2.17          & & \textbf{83.77(5.75)}\tnote{2} & 3.28(0.35)        \\
            $\text{LayoutLM}_{\text{BASE}}$ + \textbf{\vilai}(Text Block)  &  \textbf{93.38} & \textbf{0.53} &  & 92.00          & \textbf{2.10} & & 83.44(6.48)          & \textbf{2.83(0.34)}  \\
            
            \bottomrule
        \end{tabular}
        \begin{tablenotes}
            \item[1] For \dataset, we show averaged scores with standard deviation in parentheses across the 5-fold cross validation subsets. 
            \item[2] In this table, we report \dataset results using VILA structures detected by visual layout models. When the ground-truth VILA structures are available, both \vilai and \vilah models can achieve better accuracy, shown in Table \ref{table:res-s2hard-block-gen}.
        \end{tablenotes}
    \end{threeparttable}
    }
    \caption{Performance of baseline and \vilai models on the scientific document extraction task. \vilai provides consistent accuracy improvements over the baseline LayoutLM model on all three benchmark datasets.}
    \label{table:res-ivila}
    \vspace{1mm}
\end{table*}

\section{Experimental Setup}

\subsection{Implementation Details}

Our models are implemented using PyTorch~\cite{pytroch} and the transformers library~\cite{wolf-etal-2020-transformers}. A series of baseline and VILA models are fine-tuned on 4-GPU RTX8000 or A100 machines. 
The AdamW optimizer~\cite{kingma2014adam, Loshchilov2019DecoupledWD} is adopted with a $5\times10^{-5}$ learning rate and $(\beta_1,\beta_2)=(0.9,0.999)$. 
The learning rate is linearly warmed up over 5\% steps then linearly decayed.
For all datasets (GROTOAP2, DocBank, \dataset), unless otherwise specified, we select the best fine-tuning batch size (40, 40 and 12) and training epochs (24, 6,\footnote{We try to keep gradient update steps the same for the GROTOAP2 and the DocBank dataset. As DocBank contains 4$\times$ examples, the number of DocBank models' training epochs is reduced by 75\%.} and 10) for all models.
As for \dataset, given its smaller size, we use 5-fold cross validation and report averaged scores, and use $2\times10^{-5}$ learning rate with 20 epochs.
We split \dataset based on papers rather than pages to avoid exposing paper templates of test samples in the training data. 
Mixed precision training~\cite{Micikevicius2018MixedPT} is used to speed up the training process.

For \vilai models, we fine-tune several BERT-variants with VILA-enhanced text inputs, and the models are initialized from pre-trained weights available in the transformers library.
The \vilah models are initialized as mentioned in Section~\ref{sec:vila-hie}, and by default, positional information is injected for each group.

\subsection{Competing Methods} 

We consider three approaches that compete with the proposed methods from different perspectives:
\begin{enumerate}[noitemsep]
  \item \textbf{Baselines} The LayoutLM~\cite{xu2020layoutlm} model is the main baseline method. 
  It is the closest model counterpart to our VILA-augmented models as it also injects layout information and achieves previous SOTA performance on the Scientific PDF parsing task~\cite{li2020docbank}.
  \item \textbf{Sentence Breaks} For \vilai models, besides using VILA-based indicators, we also compare with indicators generated from sentence breaks detected by PySBD~\cite{sadvilkar-neumann-2020-pysbd}. 
  Figure~\ref{fig:ivila}(a) shows that the inserted sentence-break indicators may have both "false-positive" or "false-negative" hints for token semantic category changes, making it less helpful for the end task. 
  \item \textbf{Simple Group Classifier} For hierarchical models, we consider another baseline approach, where the group texts are separately fed into a LayoutLM-based group classifier.
  It doesn't require complicated model design, and uses a full LayoutLM to model each group's text, as opposed to the $l_\text{g}=1$ layer used in the \vilah models.
  However, this method cannot account for inter-group interactions, and is far less efficient.\footnote{Despite the group texts being relatively short, this method causes extra computational overhead as the full LayoutLM model needs to be run $m$ times for all groups in a page. The simple group classifier models are only trained for 5, 2, and 5 epochs for GROTOAP2, DocBank, and \dataset for tractability.}
\end{enumerate}

\subsection{Metrics}

\paragraph{Prediction Accuracy} The token label distribution is heavily skewed towards categories corresponding to paper body texts (e.g., the ``BODY\_CONTENT'' category in GROTOAP2 or the ``paragraph'' category in \dataset and DocBank). Therefore, we choose to use \mfscore as our primary evaluation metric for prediction accuracy. 

\paragraph{Group Category Inconsistency} To better characterize how different models behave with respect to group structure, we also report a diagnostic metric that calculates the uniformity of the token categories within a group. 
Hypothetically, tokens $T^{(j)}$ in the $j$-th group $g_j$ share the same category $c$, and naturally the group inherits the semantic label $c$. 
We use the group token category entropy to measure the inconsistency of a model's predicted token categories within the same group: 
\begin{equation}
  \mathrm{H}(g) = -\sum_c p_{c}\log{p_{c}}, 
\end{equation}
\noindent where $p_c$ denotes the probability of a token in group $g$ being classified as category $c$. 
When all tokens in a group have the same category, the group token category inconsistency is zero. 
$\mathrm{H}(g)$ reaches the maximum when $p_{c}$ is a uniform distribution across all possible categories. 
The inconsistency for $G$ is the arithmetic mean of all individual groups $g_i$: 
\begin{equation}
    \mathrm{H}(G) = \frac{1}{m} \sum_i^m \mathrm{H}(g_i)
\end{equation}
$\mathrm{H}(G)$ acts as an auxiliary metric for evaluating prediction quality with respect to the provided VILA structures.
In the remainder of this paper, we report the inconsistency metric for text blocks $G^{(\mathcal{B})}$ by default, and scale the values by a factor of 100. 

\paragraph{Measuring Efficiency} We report the inference time per sample as a measure of model efficiency. 
We select 1,000 pages from the GROTOAP2 test set, and report the average model runtime for 3 runs on this subset. 
All models are tested on an isolated machine with a single V100 GPU. 
We report the time incurred for text classification; time costs associated with PDF-to-text conversion or VILA structure detection are not included (these are treated as pre-processing steps, which can be cached and re-used when processing documents with different content extractors).

\begin{table*}[htbp]
    \centering
    \resizebox{\linewidth}{!}{
    \begin{threeparttable}
        \renewcommand{\arraystretch}{1.35}%

        \begin{tabular}{@{}rrrrcrrcrrcrrr@{}}
            \toprule
            & \multicolumn{2}{c}{GROTOAP2} & \phantom{\small{1}} & \multicolumn{2}{c}{DocBank}  & \phantom{\small{1}} & \multicolumn{2}{c}{\dataset}  \\
            \cmidrule{2-3} \cmidrule{5-6} \cmidrule{8-9}
            & \mfscore $\upuparrows$ & $\mathrm{H}(G)$ $\downdownarrows$ &  & \mfscore $\upuparrows$ & $\mathrm{H}(G)$ $\downdownarrows$ & & \mfscore $\upuparrows$ & $\mathrm{H}(G)$ $\downdownarrows$  & & Inference Time \small{(ms)}\\
            \midrule

            $\text{LayoutLM}_{\text{BASE}}$       &  92.34          & 0.78          &  & 91.06          & 2.64          & & 82.69(6.04) & 4.19(0.25)     & & 52.56(0.25) \\
            Simple Group Classifier          &  92.65         & 0.00          &  & 87.01          & 0.00          & & -\tnote{1}         & -         & & 82.57(0.30) 
            \\
            \cdashline{1-12}[.4pt/1pt]
            \textbf{\vilah}(Text Line) %
            & 91.65          & 0.32          &  & 91.27          & 1.07          & & 83.69(2.92)          & 1.70(0.68)        & & 28.07(0.37)\tnote{2}\\
            \textbf{\vilah}(Text Block)  %
            & 92.37          & 0.00          &  & 87.78          & 0.00          & & 82.09(5.89)          & 0.36(0.12)        & & \textbf{16.37(0.15)} \\
            \bottomrule
        \end{tabular}
        \begin{tablenotes}
            \item[1] The simple group classifier fails to converge for one run. We do not report the results for fair comparison.
            \item[2] When reporting efficiency in other parts of the paper, we use this result because of its optimal combination of accuracy and efficiency. 
        \end{tablenotes}
    \end{threeparttable}
    }
     \caption{Content extraction performance for \vilah.  The \vilah models significantly reduce the inference time cost compared to LayoutLM, while achieving comparable accuracy on the three benchmark datasets.}
    \label{table:res-hvila}
\end{table*}

\begin{table}[t]
    \centering
    \resizebox{\linewidth}{!}{
    \begin{threeparttable}
        \renewcommand{\arraystretch}{1.35}%

        \begin{tabular}{cccc}
            \toprule
            Base Model   &   Baseline         & Text Line $G^{(\mathcal{L})}$     & Text Block $G^{(\mathcal{B})}$    \\

            \midrule
            DistilBERT         & 90.52     & 91.14            & \textbf{92.12} \\
            BERT               & 90.78     & 91.65            & \textbf{92.31} \\
            RoBERTa            & 91.64     & 92.04            & \textbf{92.52} \\            
            LayoutLM           & 92.34     & 92.37            & \textbf{93.38} \\
            \bottomrule
        \end{tabular}
    \end{threeparttable}
    }
    \caption{Content extraction performance (Macro F1 on the GROTOAP2 dataset) for \vilai using different BERT model variants.  \vilai can be applied to both standard BERT-based models and layout-aware ones, and consistently improves the classification accuracy.}
    \label{table:res-vila-ind-berts}
\end{table}

\begin{figure}[t]
    \centering
    \includegraphics[width=\linewidth]{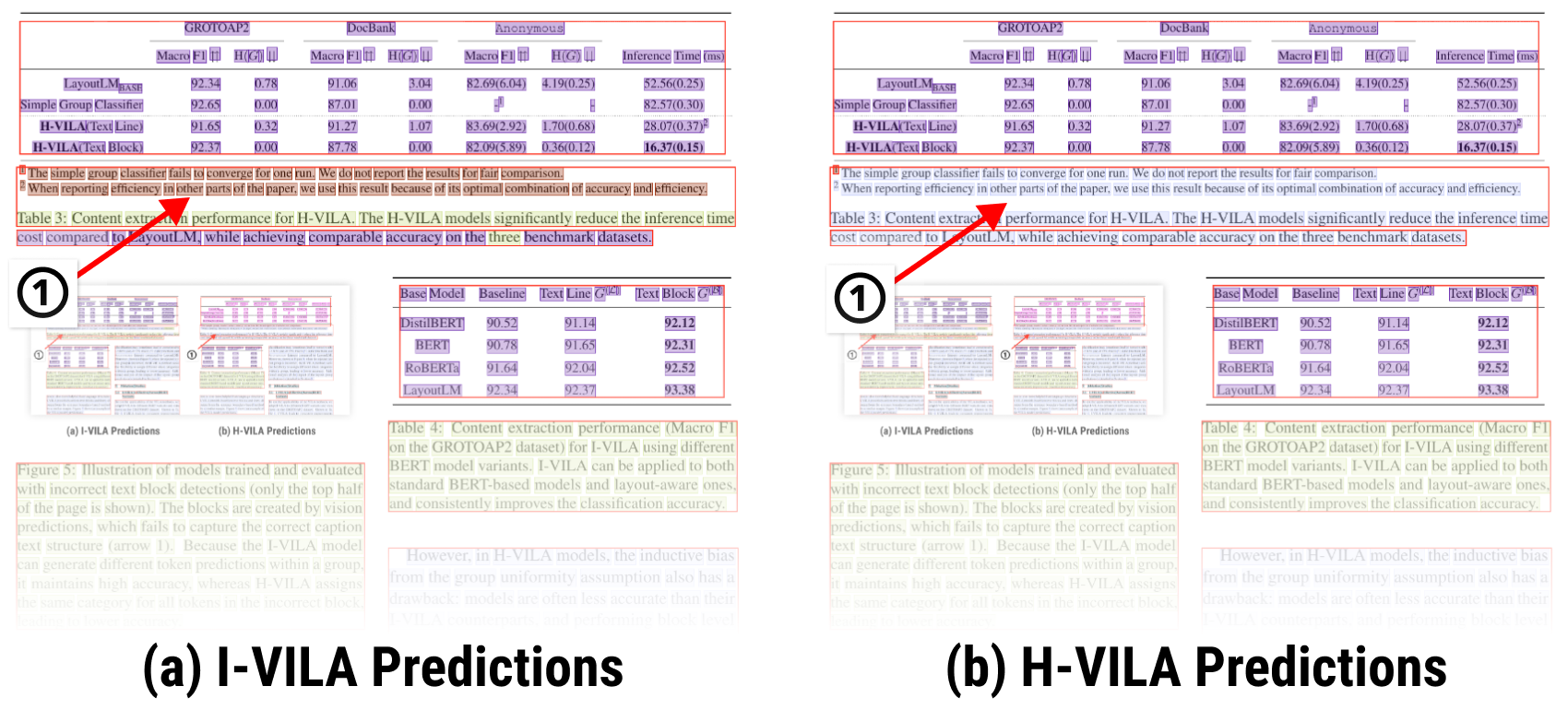}
    \caption{Illustration of models trained and evaluated with incorrect text block detections (only the top half of the page is shown). The blocks are created by vision predictions, which fails to capture the correct caption text structure (arrow 1). Because the \vilai model can generate different token predictions within a group, it maintains high accuracy, whereas \vilah assigns the same category for all tokens in the incorrect block, leading to lower accuracy. 
    } 
    \label{fig:pred-bad-block}
    \vspace{-2mm}
  \end{figure}

\section{Results}

\subsection{\vilai Achieves Better Accuracy}

Table~\ref{table:res-ivila} shows that \vilai models lead to consistent accuracy improvements without further pretraining. 
Compared to the baseline LayoutLM model, inserting layout indicators 
results in +1.13\%, +1.90\%, and +1.29\% \mfscore improvements across the three benchmark datasets. 
\vilai models also achieve better token prediction consistency; the corresponding group category inconsistency is reduced by 32.1\%, 21.7\%, and 21.7\% compared to baseline.
Moreover, VILA information is also more helpful than language structures: \vilai models based on text blocks and lines all outperform the sentence boundary-based method by a similar margin. 
Figure~\ref{fig:pred} shows an example of the VILA model predictions.

\subsection{\vilah is More Efficient}
\begin{table*}[htbp]
    \centering
    \resizebox{1\linewidth}{!}{
    \begin{threeparttable}
        \renewcommand{\arraystretch}{1.35}%

        \begin{tabular}{@{}rrrrcrrcrrcrrc@{}}
            \toprule
            & \multicolumn{2}{c}{GROTOAP2} & \phantom{\small{1}} & \multicolumn{2}{c}{DocBank}  & \phantom{\small{1}} & \multicolumn{2}{c}{\dataset}  \\
            \cmidrule{2-3} \cmidrule{5-6} \cmidrule{8-9}
            & F1 $\upuparrows$ & $\mathrm{H}(G)$ $\downdownarrows$ &  & F1 $\upuparrows$ & $\mathrm{H}(G)$ $\downdownarrows$ & & F1 $\upuparrows$ & $\mathrm{H}(G)$ $\downdownarrows$ & & Training Cost\tnote{1} \\
            \midrule
            $\text{BERT}_{\text{BASE}}$~\cite{devlin-etal-2019-bert}       &  90.78          & 1.58          &  & 87.24          & 3.50          & & 78.34(6.53)\tnote{1} & 7.17(0.95)   & & 40 hr fine-tuning   \\
            $\text{BERT}_{\text{BASE}}$ + \textbf{\vilai}(Text Line)    &  91.65          & 1.13         &  & 90.25 & 2.56          & & 81.15(4.83) & 4.76(1.28)     & & 40 hr fine-tuning    \\
            $\text{BERT}_{\text{BASE}}$ + \textbf{\vilai}(Text Block)  &  92.31 & 0.63 &  & 89.49          & 2.25 & & 81.82(4.88)         & 3.65(0.26) & & 40 hr fine-tuning \\
            \cdashline{1-12}[.4pt/1pt]
            $\text{LayoutLM}_{\text{BASE}}$~\cite{xu2020layoutlm}        &  92.34          & 0.78          &  & 91.06          & 2.64          & & 82.69(6.04) & 4.19(0.25)  & & \makecell{1.2k hr pretraining\\+ 50 hr fine-tuning}   \\
            $\text{LayoutLMv2}_{\text{BASE}}$~\cite{xu2020layoutlmv2}      &  -\tnote{2}          & -            &  & 93.33          & 1.93        & &  	83.05(4.51) & 3.34(0.82)  & & \makecell{9.6k hr pretraining\tnote{3}\\+ 130 hr fine-tuning}        \\
            \bottomrule
        \end{tabular}
        \begin{tablenotes}
            \item[1] We report the equivalent V100 GPU hours on the GROTOAP dataset in this column.  
            \item[2] LayoutLMv2 cannot be trained on the GROTOAP2 dataset because almost 30\% of its instances do not have compatible PDF images. 
            \item[3] The authors do not report the exact cost in the paper. The number is a rough estimate based on our experimental results. 
        \end{tablenotes}
    \end{threeparttable}
    }
    \caption{Comparison between \vilai models and other layout-aware methods that require expensive pretraining. \vilai achieves comparable accuracy with less than 5\% of the training cost.}
    \label{table:res-ivila-pretrianing}
\end{table*}

Table~\ref{table:res-hvila} summarizes the efficiency improvements of the \vilah models with $l_{\text{g}}=1$ and $l_{\text{p}}=12$.   
As block-level models perform predictions directly at the text block level, the group category inconsistency is naturally zero. 
Compared to LayoutLM, \vilah models with text lines brings a 46.59\% reduction in inference time, without heavily penalizing the final prediction accuracies (-0.75\%, +0.23\%, +1.21\% \mfscore).
When text blocks are used, \vilah models are even more efficient (68.85\% and 80.17\% inference time reduction compared to the LayoutLM and simple group classifier baseline), and they also achieve similar or better accuracy compared to the simple group classifier (-0.30\%, +0.88\% \mfscore for GROTOAP2 and DocBank). 

However, in \vilah models, the inductive bias from the group uniformity assumption also has a drawback: models are often less accurate than their \vilai counterparts, and performing block level classification may sometimes lead to worse results (-3.60\% and -0.73\% \mfscore in the DocBank and \dataset datasets compared to LayoutLM). 
Moreover, shown in Figure~\ref{fig:pred-bad-block}, when the injected layout group is incorrect, the \vilah method lacks the flexibility to assign different token categories within a group, leading to lower accuracy. 
Additional analysis of the impact of the layout group predictions is detailed in Section~\ref{sec:discussion}.

\section{Ablation Studies}
\label{sec:ablations}

\subsection{\vilai is Effective Across BERT Variants}
\label{sec:res-vila-ind-berts}

To test the applicability of the VILA methods, we adapt \vilai to different BERT variants and train them on the GROTOAP2 dataset.
Shown in Table~\ref{table:res-vila-ind-berts}, \vilai leads to consistent improvements on DistilBERT~\cite{sanh2019distilbert}, BERT, and RoBERTa~\cite{liu2019roberta},\footnote{Positional embeddings are not used in these models.} leading to up-to +1.77\%, +1.69\%, and 0.96\% \mfscore compared to non-VILA counterparts.

\subsection{\vilai Improves Accuracy without Pretraining}

In Table~\ref{table:res-ivila-pretrianing}, we fine-tune a series of \vilai models based on BERT, and compare their performance with LayoutLM and LayoutLMv2~\cite{xu2020layoutlmv2} which require additional large-scale pretraining on corpora with layout.   
BERT+\vilai achieves comparable accuracy to LayoutLM  (0.00\%, -0.89\%, -1.05\%), with only 5\% of the training cost.\footnote{It takes 10.5 hours to finish fine-tuning \vilai on the GROTOAP2 dataset using a 4 RTX 8000 machine, equivalent to around 60 V100 GPU hours, approximately 5\% of the 1280 hours of the pretraining time for LayoutLM.} 
\vilai also closes the gap with the latest multimodal method LayoutLMv2~\cite{xu2020layoutlmv2} with only 1\% of the training cost.
This further verifies that injecting layout indicator tokens is a novel and effective way of incorporating layout information into language models.

\begin{table*}[t]
    \centering
    \resizebox{0.9\linewidth}{!}{
    \begin{threeparttable}
        \renewcommand{\arraystretch}{1.35}%

        \begin{tabular}{rrrrcrrcrr}
            \toprule
            \multicolumn{2}{l}{} & \multicolumn{2}{c}{\textbf{Group-uniform Oracle}} & \phantom{\small{1}} & \multicolumn{2}{c}{\textbf{\vilai}} & \phantom{\small{1}} & \multicolumn{2}{c}{\textbf{\vilah}} \\
            \cmidrule{3-4} \cmidrule{6-7} \cmidrule{9-10} 
            Experiment & Group Source & Max \mfscore & $\mathrm{H}(G)$ &  & \mfscore       & $\mathrm{H}(G)$ &  & \mfscore          & $\mathrm{H}(G)$          \\
            \midrule
            \multirowcell{3}{\textbf{Varying} $G^{\mathcal{B}}$}
            & Ground-Truth  &  100.00(0.00)   &  0.00(0.00)   & & 86.50(4.52)        & 1.86(0.29)      &  & 85.91(3.13)       & 0.35(0.19)       \\
            & Vision Model &  99.31(0.23)    &  1.09(0.30) & & 83.44(6.48)	& 2.83(0.34) & & 82.09(5.89) & 0.36(0.12) \\
            & PDF Parsing  &  96.91(1.09)    &  2.06(0.86) & & 83.95(4.45)	& 3.93(0.93) & & 78.69(4.90) & 0.02(0.01) \\
            \midrule
            \multirowcell{2}{\textbf{Varying} $G^{\mathcal{L}}$}
            & Vision Model  & 99.57(0.13)    &  0.42(0.18)\tnote{1} & & 83.77(5.75)	& 1.20(0.16) & & 83.69(2.92) & 0.20(0.12) \\
            & PDF Parsing   & 99.70(0.12)    &  0.38(0.26) & & 82.97(5.56)	& 1.28(0.13) & & 82.61(4.10) & 0.00(0.00) \\
            \bottomrule
        \end{tabular}    
        \begin{tablenotes}
            \item[1] For text line detector experiments, we report $\mathrm{H}(G)$ based on text lines rather than blocks. 
        \end{tablenotes}
    \end{threeparttable}
    }
    \caption{VILA model performance when using different layout group detectors for text blocks $G^{(\mathcal{B})}$ and lines $G^{(\mathcal{L})}$ on the \dataset dataset.}
    \label{table:res-s2hard-block-gen}
\end{table*}

\section{VILA in Practice: The Impact of Layout Group Detectors}
\label{sec:discussion}

Applying VILA methods in practice requires running a group layout detector as a critical first step.  In this section, we analyze how the accuracy of different block and line group detectors affects the accuracy of \vilah and \vilai models.

The results are shown in Table~\ref{table:res-s2hard-block-gen}. 
We report on the \dataset dataset using two automated group detectors: the CERMINE PDF parser~\cite{tkaczyk2015cermine} and the Mask R-CNN vision model trained on the PubLayNet dataset~\cite{zhong2019publaynet}.  We also report on using ground truth blocks as an upper bound.  The ``Group-uniform Oracle'' illustrates how well the different group detectors reflect the group uniformity assumption; 
in the oracle setting, one is given ground truth labels but is restricted to assigning the same label to all tokens in a group.

When using text blocks, the performance of \vilah hinges on the accuracy of group detection, while \vilai shows more reliable results when using different group detectors. 
This suggests that improvements in vision models for block detection could be a promising avenue for improving content extraction performance, especially when using \vilah, and
\vilai may be the better choice when block detection accuracy is lower.

We also observe that text line-based methods tend to be higher performing for both group detectors, by a small margin for \vilai and a larger one for \vilah.  The group detectors in our experiments are trained on data from PubLayNet, and applied to a different dataset, \dataset.  This domain transfer affects block detectors more than line detectors, because the two datasets define blocks differently.  This setting is realistic because ground truth blocks from the target dataset may not always be available for training (even when labeled tokens are). Training a group detector on \dataset is likely to improve performance.

\section{Conclusion}

In this paper, we introduce two new ways to integrate Visual Layout (VILA) structures into the NLP pipeline for structured content extraction from scientific paper PDFs. 
We show that inserting special indicator tokens based on VILA (\vilai) can lead to robust improvements in token classification accuracy (up to +1.9\% \mfscore) and consistency (up to -32\% group category inconsistency). 
In addition, we design a hierarchical transformer model based on VILA (\vilah), which can reduce inference time by 46\% with less than 0.8\% \mfscore reduction compared to previous SOTA methods. 
These VILA-based methods can be easily incorporated into different BERT variants with only fine-tuning, achieving comparable performance against existing work
with only 5\% of the training cost.
We ablate the influence of different visual layout detectors on VILA-based models, and provide suggestions for practical use. 
We release a benchmark suite, along with a newly curated dataset \dataset, to systematically evaluate the proposed methods.

Our study is well-aligned with the recent exploration of injecting structures into language models, and provides new perspectives on how to incorporate documents' visual structures.
The approach shows how explicitly modeling task structure can help achieve "green AI" goals, dramatically reducing computation and energy costs without significant loss in accuracy.
While we evaluate on scientific documents, related visual group structures also exist in other kinds of documents, and adapting our techniques to those domains could offer improvements in corporate reports, historical archives, or legal documents, and this is an item of future work.

\section*{Acknowledgements}
We thank the anonymous reviewers and TACL editors for their comments and feedback on our draft, and we thank Ruochen Zhang, Mark Neumann, Rodney Kinney, Dirk Groeneveld, and Mike Cafarella for the helpful discussion and suggestions. This project is supported in part by NSF Grant OIA-2033558, NSF RAPID award 2040196, ONR grant N00014-21-1-2707, and the University of Washington WRF/Cable Professorship.

\bibliography{anthology,custom}
\bibliographystyle{acl_natbib}

\cleardoublepage
\appendix

\begin{table*}[]
    \resizebox{\linewidth}{!}{
    \begin{threeparttable}
        \renewcommand{\arraystretch}{1.25}
        \begin{tabular}{@{}lcccccccc@{}}
            \toprule
                                                                          & \textbf{Abstract}                                              & \textbf{Acknowledgment} & \textbf{Affiliation} & \textbf{Author} & \textbf{\begin{tabular}[c]{@{}c@{}}Author\\ Title\end{tabular}} & \textbf{\begin{tabular}[c]{@{}c@{}}Bib\\ Info\end{tabular}} & \textbf{\begin{tabular}[c]{@{}c@{}}Body\\ Content\end{tabular}} & \textbf{\begin{tabular}[c]{@{}c@{}}Conflict\\ Statement\end{tabular}} \\ \midrule
            $\text{BERT}_{\text{BASE}}$                                   & 97.42                                                          & 95.83                   & 96.12                & 96.91           & 96.09                                                           & 95.00                                                       & 98.80                                                           & 88.66                                                                 \\
            $\text{BERT}_{\text{BASE}}$ + \textbf{\vilai}(Text Line)      & 97.65                                                          & 95.89                   & 96.61                & 97.17           & 96.48                                                           & 95.78                                                       & 98.93                                                           & 88.28                                                                 \\
            $\text{BERT}_{\text{BASE}}$ + \textbf{\vilai}(Text Block)     & 97.67                                                          & 96.46                   & 96.80                & 97.23           & 97.73                                                           & 96.29                                                       & 98.99                                                           & 91.88                                                                 \\
            \cdashline{1-9}[.4pt/1pt]
            $\text{LayoutLM}_{\text{BASE}}$                               & 98.05                                                          & 96.29                   & 96.64                & \textbf{97.49}  & 96.51                                                           & 96.74                                                       & 99.06                                                           & 91.16                                                                 \\
            $\text{LayoutLM}_{\text{BASE}}$ + Sentence Breaks             & 97.92                                                          & 96.32                   & 96.68                & 96.74           & 95.42                                                           & 96.77                                                       & 99.11                                                           & 90.42                                                                 \\
            $\text{LayoutLM}_{\text{BASE}}$ + \textbf{\vilai}(Text Line)  & 97.99                                                          & 96.41                   & 96.72                & 97.29           & 95.98                                                           & 96.66                                                       & 99.11                                                           & 90.75                                                                 \\
            $\text{LayoutLM}_{\text{BASE}}$ + \textbf{\vilai}(Text Block) & 98.12                                                          & \textbf{96.81}          & 96.93                & 96.96           & 97.52                                                           & \textbf{96.87}                                              & 99.14                                                           & 91.43                                                                 \\
            \cdashline{1-9}[.4pt/1pt]
            Simple Group Classifier                                       & 96.10                                                          & 95.53                   & \textbf{97.10}       & 97.48           & \textbf{97.94}                                                  & 96.68                                                       & 98.94                                                           & \textbf{93.25}                                                        \\
            \textbf{\vilah}(Text Line)                                    & \textbf{98.47}                                                 & 95.88                   & 96.21                & 97.46           & 95.26                                                           & 96.68                                                       & \textbf{99.16}                                                  & 89.67                                                                 \\
            \textbf{\vilah}(Text Block)                                   & 98.01                                                          & 96.45                   & 96.14                & 97.38           & 96.31                                                           & 96.33                                                       & 99.08                                                           & 91.67                                                                 \\
            \midrule
            \# Tokens in Class                                            & 395788                                                         & 88531                   & 90775                & 26742           & 7083                                                            & 223739                                                      & 7567934                                                         & 22289                                                                 \\
            \bottomrule \\[0.125cm]  
            \toprule
            \emph{contd.}                                                 & \textbf{Copyright}                                             & \textbf{Correspondence} & \textbf{Dates}       & \textbf{Editor} & \textbf{Equation}                                               & \textbf{Figure}                                             & \textbf{Glossary}                                               & \textbf{Keywords}                                                     \\
            \midrule
            $\text{BERT}_{\text{BASE}}$                                   & 97.34                                                          & 89.66                   & 94.56                & 99.71           & 17.60                                                           & 94.05                                                       & 80.18                                                           & 93.42                                                                 \\
            $\text{BERT}_{\text{BASE}}$ + \textbf{\vilai}(Text Line)      & 97.38                                                          & 89.57                   & 94.60                & 99.93           & 25.00                                                           & 94.84                                                       & 81.35                                                           & 94.34                                                                 \\
            $\text{BERT}_{\text{BASE}}$ + \textbf{\vilai}(Text Block)     & 97.85                                                          & 91.29                   & 94.99                & 99.95           & 29.46                                                           & 95.52                                                       & 80.45                                                           & 95.40                                                                 \\
            \cdashline{1-9}[.4pt/1pt]
            $\text{LayoutLM}_{\text{BASE}}$                               & 97.63                                                          & 89.99                   & 94.80                & 99.90           & 30.78                                                           & 95.52                                                       & 83.83                                                           & 94.95                                                                 \\
            $\text{LayoutLM}_{\text{BASE}}$ + Sentence Breaks             & 97.62                                                          & 90.07                   & 94.73                & 99.95           & 20.73                                                           & 95.83                                                       & 84.99                                                           & 93.88                                                                 \\
            $\text{LayoutLM}_{\text{BASE}}$ + \textbf{\vilai}(Text Line)  & 97.47                                                          & 90.97                   & 95.20                & 99.93           & 26.42                                                           & 95.67                                                       & 84.16                                                           & 94.82                                                                 \\
            $\text{LayoutLM}_{\text{BASE}}$ + \textbf{\vilai}(Text Block) & 97.66                                                          & 91.04                   & 95.13                & \textbf{100.00} & \textbf{39.28}                                                  & 95.74                                                       & \textbf{87.00}                                                  & \textbf{96.23}                                                        \\
            \cdashline{1-9}[.4pt/1pt]
            Simple Group Classifier                                       & 97.56                                                          & \textbf{92.11}          & \textbf{95.47}       & \textbf{100.00} & 33.17                                                           & 95.77                                                       & 80.35                                                           & 95.64                                                                 \\
            \textbf{\vilah}(Text Line)                                    & 97.78                                                          & 89.96                   & 94.98                & 99.91           & 15.60                                                           & 95.63                                                       & 84.01                                                           & 93.69                                                                 \\
            \textbf{\vilah}(Text Block)                                   & \textbf{97.98}                                                 & 90.37                   & 94.92                & 100.00          & 30.64                                                           & \textbf{95.86}                                              & 78.29                                                           & 96.15                                                                 \\
            \midrule
            \# Tokens in Class                                            & 57419                                                          & 26653                   & 23702                & 2937            & 761                                                             & 581554                                                      & 2807                                                            & 7012                                                                  \\
            \bottomrule \\[0.125cm] 
            \toprule
            \emph{contd.}                                                 & \textbf{\begin{tabular}[c]{@{}c@{}}Page\\ Number\end{tabular}} & \textbf{References}     & \textbf{Table}       & \textbf{Title}  & \textbf{Type}                                                   & \textbf{Unknown}                                            & \multicolumn{1}{l}{}                                            & \textbf{\mfscore}                                                     \\
            \midrule
            $\text{BERT}_{\text{BASE}}$                                   & 98.32                                                          & 99.60                   & 94.11                & 97.60           & 87.62                                                           & 88.60                                                       & \multicolumn{1}{l}{}                                            & 90.78                                                                 \\
            $\text{BERT}_{\text{BASE}}$ + \textbf{\vilai}(Text Line)      & 98.82                                                          & 99.60                   & 94.53                & 97.77           & 93.70                                                           & 88.14                                                       & \multicolumn{1}{l}{}                                            & 91.65                                                                 \\
            $\text{BERT}_{\text{BASE}}$ + \textbf{\vilai}(Text Block)     & 98.92                                                          & 99.64                   & 94.31                & 98.19           & 93.09                                                           & 88.81                                                       & \multicolumn{1}{l}{}                                            & 92.31                                                                 \\
            \cdashline{1-9}[.4pt/1pt]
            $\text{LayoutLM}_{\text{BASE}}$                               & 98.94                                                          & 99.62                   & 95.30                & 97.91           & 91.24                                                           & 89.19                                                       & \multicolumn{1}{l}{}                                            & 92.34                                                                 \\
            $\text{LayoutLM}_{\text{BASE}}$ + Sentence Breaks             & 98.90                                                          & 99.61                   & 95.63                & 98.13           & 91.68                                                           & 89.14                                                       & \multicolumn{1}{l}{}                                            & 91.83                                                                 \\
            $\text{LayoutLM}_{\text{BASE}}$ + \textbf{\vilai}(Text Line)  & \textbf{99.05}                                                 & 99.63                   & 95.61                & 97.80           & 94.59                                                           & 89.86                                                       & \multicolumn{1}{l}{}                                            & 92.37                                                                 \\
            $\text{LayoutLM}_{\text{BASE}}$ + \textbf{\vilai}(Text Block) & \textbf{99.05}                                                 & 99.65                   & 95.73                & \textbf{98.39}  & \textbf{95.17}                                                  & \textbf{90.47}                                              & \multicolumn{1}{l}{}                                            & \textbf{93.38}                                                        \\
            \cdashline{1-9}[.4pt/1pt]
            Simple Group Classifier                                       & 99.02                                                          & 99.61                   & 93.94                & 98.18           & 94.91                                                           & 89.60                                                       & \multicolumn{1}{l}{}                                            & 92.65                                                                 \\
            \textbf{\vilah}(Text Line)                                    & 98.96                                                          & 99.63                   & \textbf{96.02}       & 97.76           & 93.61                                                           & 90.00                                                       & \multicolumn{1}{l}{}                                            & 91.65                                                                 \\
            \textbf{\vilah}(Text Block)                                   & 99.16                                                          & \textbf{99.68}          & 95.00                & 98.36           & 95.07                                                           & 89.23                                                       & \multicolumn{1}{l}{}                                            & 92.37                                                                 \\
            \midrule
            \# Tokens in Class                                            & 46884                                                          & 2340796                 & 558103               & 22110           & 4543                                                            & 54639                                                       & \multicolumn{1}{l}{}                                            & ---                                                                   \\ \bottomrule
            \end{tabular}
    \end{threeparttable}
    }
    \caption{Prediction F1 breakdown for all models on the GROTOAP2 dataset.}
    \label{table:perclass-grotoap2}
\end{table*}
\begin{table*}[]
    \resizebox{1.\linewidth}{!}{
    \begin{threeparttable}
        \renewcommand{\arraystretch}{1.25}
        \begin{tabular}{@{}lccccccccccccc@{}}
            \toprule
                                                                          & \textbf{Abstract} & \textbf{Author} & \textbf{Caption} & \textbf{Date} & \textbf{Figure} & \textbf{Footer} & \textbf{List} & \textbf{Paragraph} & \textbf{Reference} & \textbf{Section} & \textbf{Table} & \textbf{Title} & \mfscore \\ \midrule
            {\small $\text{BERT}_{\text{BASE}}$}                                   & 97.82             & 89.96           & 93.91            & 87.33          & 71.97           & 84.76           & 75.99          & 96.84              & 92.05              & 92.81            & 74.19          & 89.31          & 87.24          \\
            {\small $\text{BERT}_{\text{BASE}}$ + \textbf{\vilai}(Text Line)}      & 97.99             & 90.67           & 95.74            & 88.12          & 88.85           & 88.29           & 80.20          & 97.85              & 92.68              & 94.91            & 77.39          & 90.34          & 90.25          \\
            {\small $\text{BERT}_{\text{BASE}}$ + \textbf{\vilai}(Text Block)}     & 98.15             & 90.66           & 96.56            & 87.83          & 79.49           & 88.40           & 80.72          & 97.51              & 92.62              & 94.86            & 76.91          & 90.22          & 89.49          \\
            \cdashline{1-14}[.4pt/1pt]
            {\small $\text{LayoutLM}_{\text{BASE}}$}                               & 98.63             & 92.25           & 96.88            & 87.13          & 76.56           & 94.26           & 89.67          & 97.72              & 93.16              & 96.31            & 77.38          & 92.80          & 91.06          \\
            {\small $\text{LayoutLM}_{\text{BASE}}$ + Sentence Breaks}             & 98.48             & 92.70           & 96.93            & 88.06          & 77.65           & 94.35           & 90.46          & 97.81              & 92.61              & 96.58            & 78.84          & 92.81          & 91.44          \\
            {\small $\text{LayoutLM}_{\text{BASE}}$ + \textbf{\vilai}(Text Line)}  & 98.57             & 92.64           & 97.35            & 87.87          & \textbf{90.78}  & 94.37           & 90.77          & 98.44              & 92.87              & \textbf{96.60}   & 80.43          & 92.78          & 92.79          \\
            {\small $\text{LayoutLM}_{\text{BASE}}$ + \textbf{\vilai}(Text Block)} & \textbf{98.68}    & 92.31           & 97.44            & 87.69          & 83.41           & 94.03           & 90.56          & 98.13              & 93.27              & 96.44            & 79.51          & 92.48          & 92.00          \\
            {\small $\text{LayoutLMv2}_{\text{BASE}}$}                             & \textbf{98.68}    & \textbf{93.04}  & \textbf{97.49}   & \textbf{89.55} & 85.60           & \textbf{95.30}  & \textbf{93.63} & \textbf{98.46}     & \textbf{94.30}     & 96.48            & \textbf{84.41} & 93.10          & \textbf{93.34} \\
            \cdashline{1-14}[.4pt/1pt]
            {\small Simple Group Classifier}                                       & 93.85             & 84.68           & 96.55            & 71.04          & 80.63           & 91.58           & 83.84          & 97.53              & 92.54              & 85.33            & 73.85          & 92.65          & 87.01          \\
            {\small \textbf{\vilah}(Text Line)}                                    & \textbf{98.68}    & 90.95           & 95.46            & 80.99          & 88.79           & 93.84           & 90.77          & 98.36              & 93.81              & 95.27            & 78.46          & 89.81          & 91.27          \\
            {\small \textbf{\vilah}(Text Block)}                                   & 98.57             & 86.81           & 95.76            & 70.33          & 80.29           & 91.23           & 79.82          & 97.53              & 92.97              & 86.70            & 79.84          & \textbf{93.52} & 87.78          \\
            \midrule
            \# Tokens in Class                                            & 461898            & 81061           & 858862           & 3275          & 932150          & 158176          & 684786        & 20630188           & 1813594            & 154062           & 235801         & 26355          & ---      \\ \bottomrule
            \end{tabular}
    \end{threeparttable}
    }
\caption{Prediction F1 breakdown for all models on the DocBank dataset.}
\label{table:perclass-docbank}
\end{table*}

\begin{table*}[]
    \resizebox{\linewidth}{!}{
    \begin{threeparttable}
        \renewcommand{\arraystretch}{1.25}
        \begin{tabular}{@{}lcccccccc@{}}
            \toprule
                                     & \textbf{Abstract} & \textbf{Author}   & \textbf{Bibliography} & \textbf{Caption}   & \textbf{Equation} & \textbf{Figure} & \textbf{Footer}                     & \textbf{Footnote}      \\ \midrule
            $\text{BERT}_{\text{BASE}}$                                   & 91.67(5.51)       & \textbf{71.38(18.79)}      & 97.90(1.59)           & 94.64(1.38)        & 76.23(4.36)       & 60.14(24.13)    & 61.99(17.04)                        & 62.91(7.23)            \\
            $\text{BERT}_{\text{BASE}}$ + \textbf{\vilai}(Text Line)      & 89.38(6.50)       & 65.93(15.48)      & 97.92(1.56)           & 96.66(1.39)        & 83.22(5.87)       & 72.11(13.35)    & 57.75(22.46)                        & 72.78(12.45)           \\
            $\text{BERT}_{\text{BASE}}$ + \textbf{\vilai}(Text Block)     & 90.45(3.61)       & 64.97(16.11)      & 97.21(1.27)           & 96.82(0.94)        & 83.56(5.59)       & 70.57(11.56)    & 59.79(23.18)                        & 80.17(10.48)           \\
            \cdashline{1-9}[.4pt/1pt]
            $\text{LayoutLM}_{\text{BASE}}$                               & 91.87(4.89)       & 69.39(11.30)      & 98.08(1.13)           & 92.98(7.35)        & 77.49(7.13)       & 74.46(18.48)    & 67.42(18.90)                        & 77.22(17.59)           \\
            $\text{LayoutLM}_{\text{BASE}}$ + Sentence Breaks             & 92.01(4.79)       & 69.22(11.02)      & \textbf{98.57(1.24)}           & 95.74(1.36)        & 77.94(9.68)       & 67.80(25.61)    & \textbf{69.67(20.06)}                        & 78.57(16.45)           \\
            $\text{LayoutLM}_{\text{BASE}}$ + \textbf{\vilai}(Text Line)  & 91.77(5.85)       & 69.81(7.86)       & 98.09(1.64)           & 94.06(2.91)        & 84.48(7.00)       & 71.57(21.49)    & 67.23(23.01)                        & 77.10(15.64)           \\ 
            $\text{LayoutLM}_{\text{BASE}}$ + \textbf{\vilai}(Text Block) & 92.91(4.02)       & 70.42(13.38)      & 98.19(1.57)           & \textbf{97.19(1.16)}        & 83.76(6.61)       & 68.38(26.11)    & 68.03(19.11)                        & 76.77(17.64)           \\
            $\text{LayoutLMv2}_{\text{BASE}}$                             & 91.09(6.46)       & 63.42(17.55)      & 97.74(2.00)           & 96.73(1.39)        & 77.18(13.70)      & \textbf{83.71(11.53)}    & 64.37(22.24)                        & 70.20(12.43)           \\ 
            \cdashline{1-9}[.4pt/1pt]
            \textbf{\vilah}(Text Line)                                    & \textbf{93.90(5.16)}       & 70.86(9.78)       & 97.71(1.26)           & 92.86(3.89)        & 81.38(7.79)       & 77.86(10.65)    & 65.95(23.44)                        & 81.76(15.03)           \\ 
            \textbf{\vilah}(Text Block)                                   & 93.40(6.14)       & 67.03(19.43)      & 96.11(3.38)           & 92.76(6.47)        & \textbf{86.87(8.64)}       & 79.64(11.21)    & 63.72(22.01)                        & \textbf{83.66(9.88)}            \\
            \midrule
            \# Tokens in Class                                    & 2854(432)         & 543(118)          & 15681(3704)           & 4046(2119)         & 2552(1872)        & 1402(1316)      & 480(205)        & 2468(1254)             \\
            \bottomrule \\[0.25cm] 
            \toprule
            \emph{contd.}                                                 & \textbf{Header}   & \textbf{Keywords} & \textbf{List}         & \textbf{Paragraph} & \textbf{Section}  & \textbf{Table}  & \textbf{Title} & \textbf{\mfscore} \\ 
            \midrule
            $\text{BERT}_{\text{BASE}}$                                   & 76.47(8.51)       & 90.16(6.44)       & 51.00(16.90)          & 96.07(1.37)        & 79.72(3.46)       & 79.93(16.26)    & 84.81(8.52)    & 78.34(6.53)            \\
            $\text{BERT}_{\text{BASE}}$ + \textbf{\vilai}(Text Line)      & 81.53(7.94)       & 87.06(5.57)       & 58.64(8.10)           & 96.67(1.13)        & 87.21(3.25)       & 85.58(15.67)    & 84.80(5.84)    & 81.15(4.83)            \\
            $\text{BERT}_{\text{BASE}}$ + \textbf{\vilai}(Text Block)     & 83.99(8.74)       & 87.86(7.51)       & 62.01(13.25)          & 96.65(1.21)        & 86.71(3.23)       & 80.44(16.35)    & 86.14(5.23)    & 81.82(4.88)            \\
            \cdashline{1-9}[.4pt/1pt]
            $\text{LayoutLM}_{\text{BASE}}$                               & 88.21(5.81)       & 88.14(5.94)       & 58.21(15.15)          & 96.88(0.87)        & 88.14(2.73)       & 82.02(15.58)    & 89.90(8.17)    & 82.69(6.04)            \\
            $\text{LayoutLM}_{\text{BASE}}$ + Sentence Breaks             & 88.08(5.71)       & 88.80(3.23)       & 60.61(11.80)          & 97.01(0.85)        & 88.05(2.79)       & 81.59(16.22)    & 88.52(5.92)    & 82.81(5.21)            \\
            $\text{LayoutLM}_{\text{BASE}}$ + \textbf{\vilai}(Text Line)  & 87.14(6.49)       & 86.66(6.24)       & 65.82(10.92)          & \textbf{97.17(1.26)}        & \textbf{89.79(2.48)}       & \textbf{86.00(12.33)}    & 89.89(7.47)    & \textbf{83.77(5.75)}            \\
            $\text{LayoutLM}_{\text{BASE}}$ + \textbf{\vilai}(Text Block) & \textbf{88.39(6.20)}       & \textbf{90.92(3.97)}       & 59.06(17.99)          & \textbf{97.17(1.14)}        & 88.67(3.57)       & 81.84(15.77)    & \textbf{89.95(6.32)}    & 83.44(6.48)            \\
            $\text{LayoutLMv2}_{\text{BASE}}$                             & 86.95(6.84)       & 89.71(7.95)       & \textbf{68.36(10.05)}          & 96.65(0.71)        & 89.48(4.13)       & 81.69(15.05)    & 88.46(6.00)    & 83.05(4.51)            \\
            \cdashline{1-9}[.4pt/1pt]
            \textbf{\vilah}(Text Line)                                    & 87.89(6.45)       & 86.34(5.02)       & 65.76(10.26)          & 96.90(0.75)        & 85.45(2.02)       & 85.19(7.55)     & 85.62(6.00)    & 83.69(2.92)            \\ 
            \textbf{\vilah}(Text Block)                                   & 86.49(6.08)       & 76.97(18.82)      & 55.82(16.99)          & 96.43(1.40)        & 86.72(4.55)       & 81.38(14.94)    & 84.39(9.10)    & 82.09(5.89)            \\
            \midrule
            \# Tokens in Class                                    & 1122(463)         & 130(27)           & 2274(593)             & 95732(8226)        & 882(113)          & 3887(2041)      & 240(26)      & ---   \\ 
            \bottomrule
            \end{tabular}
    \end{threeparttable}
    }
    \caption{Prediction F1 breakdown for all models on the \dataset dataset. Similar to the results in the main paper, we show averaged scores with standard deviation in parentheses across the 5-fold cross validation subsets.}
    \label{table:perclass-s2hard}
\end{table*}

\section{Model Performance Breakdown}

In Table~\ref{table:perclass-grotoap2}, \ref{table:perclass-docbank}, and \ref{table:perclass-s2hard}, we present model accuracies on GROTOAP2, DocBank, and \dataset of each category for the results reported in the main paper.

\section{Improvements of the DocBank Dataset}

We implement several fixes for the public version of the DocBank dataset to improve its accuracy and create faithful VILA structures.

\subsection{Dataset Artifacts}

As the DocBank dataset is automatically generated via parsing LaTeX source from arXiv, it will inevitably include noise. 
Moreover, the authors only release the document screenshots and token information parsed using PDFMiner\footnote{\url{https://github.com/euske/pdfminer} (last accessed Jan. 1, 2022).} instead of the source PDF files, which causes additional issues when using the dataset.  
We identify some major error categories during the course of our project, detailed as follows: 

\paragraph{Incorrect PDF Parsing} 
The PDFMiner software does not work perfectly when parsing CID fonts,\footnote{\url{https://en.wikipedia.org/wiki/PostScript_fonts} (last accessed Jan. 1, 2022).} which are often used for rendering special symbols in PDFs. 
For example, the software may incorrectly parse 25\degree C as \texttt{25(cid:176) C}. 
Including such \texttt{(cid:*)} tokens in the input text is not reasonable, because they break the natural flow of the text and most pre-trained language model tokenizers cannot appropriately encode such tokens. 

\paragraph{Erroneous Label Generation}
Token labels in DocBank are extracted by parsing latex commands. 
For example, it will label all text in the  command \texttt{\textbackslash abstract\{*\}} as ``abstract''. 
Though theoretically this approach may work well for ``standard'' documents, we find the resulting label quality is far from ideal when processing real-world documents at scale. 
One major issue is that it cannot appropriately handle user-created macros, which are often used for compiling complex math equations.
It leads to very low (label) accuracy in the ``equation'' category in the dataset -- in fact, we manually inspected 10 pages, and found 60\% of the math equation tokens are wrongly labeled as other classes. This approach also fails to appropriately label some document texts that are passively generated with the LaTeX commands, e.g., the "Figure *" produced by the \texttt{\textbackslash caption} command is treated as ``paragraph''. 

\paragraph{Lack of VILA Structures} 
As the DocBank dataset generating method solely operates on the document tex sources, it does not include visual layout information. 
The missing VILA structures leads to low label accuracy for layout-sensitive categories like figure and tables -- for example, when a figure contains selectable text (i.e., it is not stored in a format like PNG or JPG, but instead contains text tokens returned by the PDF parser),
the method cannot recognize such tokens and thus it assigns incorrect labels (other than ``figure''). 
Though the authors tried to create layout group structures by applying connected component analysis method to PDF tokens,\footnote{The algorithm iteratively selects and groups adjacent tokens with the same category, and ultimately produces a list of token collections that approximate the layout groups.} we observed different types of errors in the generated groups, e.g., mis-identifying paragraph breaks (combining multiple paragraph blocks into one) or overlapping layout groups (caused by incorrect token labels), and chose not to use them. 

\subsection{Fixes and Enhancement}

Based on the aforementioned issues, we implement the following fixes and enhance the DocBank dataset with VILA structures.

\paragraph{Remove Incorrect PDF Tokens}
Provided that there are no simple ways to recover the incorrect \texttt{(cid:*)} tokens generated by PDFMiner, we simply remove them from the input text. 

\paragraph{Generate VILA Structures}
We use pre-trained Faster-RCNN models~\cite{ren2015faster} from the LayoutParser~\cite{shen2021layoutparser} tool to identify both the text lines and blocks based on the page images.  
Specifically, for text blocks, we use the \texttt{PubLayNet/mask\_rcnn\_R\_50\_FPN\_3x/} model to detect the body content regions (including title, paragraph, figure, table, and list) and the \texttt{MFD/faster\_rcnn\_R\_50\_FPN\_3x/} model to detect the display math equation regions. 
We also fine-tune a Fast RCNN model on the GROTOAP2 dataset (which has text line annotation), and use it to detect the text lines. 
All other regions (or texts that are not covered by the detected blocks or lines) are created by the connected component analysis method. 

\paragraph{Correct Label Errors}
Given the VILA structures, we can easily correct some previously mentioned errors like incorrect labels for ``Figure *'' by applying majority voting for token labels in a text block. 
However, for the ``equation'' category, given the low accuracy of the original DocBank labels, neither majority voting nor other automatic methods can easily recover the correct token categories. 
Hence, we choose to discard this category in the modeling phase, i.e., converting all existing ``equation'' labels to the background category ``paragraph''.

We update our methods for several rounds to coordinate the fixes and enhancements, and ultimately we can reduce more than 90\% of the label errors for figure and table captions.
By using the accurate pre-trained layout detection models, the generated VILA structures are more than 95\% accurate.\footnote{We randomly sample 30 pages from both the training and test dataset, and annotate the number of the incorrect text blocks for each page. A text block is considered as incorrect when it wrongly merges multiple regions (e.g., two paragraphs or one paragraph and the adjacent section header) or splits regions (e.g., generating multiple blocks for one paragraph). We report the average of page block accuracy.}

\end{document}